\documentclass[final]{cvpr}

\usepackage{times}
\usepackage{epsfig}
\usepackage{graphicx}
\usepackage{amsmath}
\usepackage{amssymb}
\usepackage{enumitem}
\usepackage{lipsum}
\usepackage[pagebackref=true,breaklinks=true,colorlinks,bookmarks=false]{hyperref}
\usepackage{algorithm} 
\usepackage{algorithmic}  
\usepackage[algo2e]{algorithm2e} 
\usepackage[para,online,flushleft]{threeparttable}
\usepackage{makecell}
\usepackage{booktabs}
\usepackage{multirow}
\def\cvprPaperID{10597} 
\def\confYear{CVPR 2021}

\begin{document}
\title{NPAS: A Compiler-aware Framework of Unified Network Pruning and Architecture Search for Beyond Real-Time Mobile Acceleration}


\author{
Zhengang Li\text{$^\star$}\textsuperscript{1},
Geng Yuan\text{$^\star$}\textsuperscript{1}, 
Wei Niu\text{$^\star$}\textsuperscript{2},  
Pu Zhao\text{$^\star$}\textsuperscript{1}\thanks{$^\star$These authors contributed equally.}, 
Yanyu Li\textsuperscript{1}, 
Yuxuan Cai\textsuperscript{1}, \\
Xuan Shen\textsuperscript{1}, 
Zheng Zhan\textsuperscript{1},
Zhenglun Kong\textsuperscript{1}, 
Qing Jin\textsuperscript{1}, 
Zhiyu Chen\textsuperscript{3}, \\
Sijia Liu\textsuperscript{4}, 
Kaiyuan Yang\textsuperscript{3}, 
Bin Ren\textsuperscript{2}, 
Yanzhi Wang\textsuperscript{1}, 
Xue Lin\textsuperscript{1}\\
\textsuperscript{1}Northeastern University, 
\textsuperscript{2}College of William and Mary,\\ 
\textsuperscript{3}Rice University, 
\textsuperscript{4}Michigan State University \\
{\tt\small \textsuperscript{\rm 1}\{li.zhen, yuan.geng, zhao.pu, li.yanyu, cai.yuxu, shen.xu, zhan.zhe, kong.zhe,} \\
{\tt\small jinqingking, yanz.wang, xue.lin\}@northeastern.edu} \\
{\tt\small \textsuperscript{\rm 2}wniu@email.wm.edu, bren@cs.wm.edu}, {\tt\small \textsuperscript{\rm 3}\{zc37, kyang\}@rice.edu}, 
{\tt\small \textsuperscript{\rm 4}liusiji5@msu.edu}
}

\maketitle


\begin{abstract}

With the increasing demand to efficiently deploy DNNs on mobile edge devices, it becomes much more important to reduce unnecessary computation and increase the execution speed. Prior methods towards this goal, including model compression and network architecture search (NAS), are largely performed independently, and do not fully consider compiler-level optimizations which is a must-do for mobile acceleration. In this work, we first propose (i) a general category of fine-grained structured pruning applicable to various DNN layers, and (ii) a comprehensive, compiler automatic code generation framework supporting different DNNs and different pruning schemes, which bridge the gap of model compression and NAS. We further propose NPAS, a compiler-aware unified network pruning and architecture search. To deal with large search space, we propose a meta-modeling procedure based on reinforcement learning with fast evaluation and Bayesian optimization, ensuring the total number of training epochs comparable with representative NAS frameworks. Our framework achieves 6.7ms, 5.9ms, and 3.9ms ImageNet inference times with 78.2\%, 75\% (MobileNet-V3 level), and 71\% (MobileNet-V2 level) Top-1 accuracy respectively on an off-the-shelf mobile phone, consistently outperforming prior work. 
\end{abstract}



\section{Introduction} 

The growing popularity of mobile AI applications and the demand for real-time Deep Neural Network (DNN) executions raise significant challenges for DNN accelerations.
However, the ever-growing size of DNN models causes intensive computation and memory cost, which impedes the deployment on resource limited mobile devices.

DNN \textbf{weight pruning} ~\cite{wen2016learning,guo2016dynamic,min20182pfpce,he2018amc,he2019filter} has been proved as an effective model compression technique that can remove redundant weights of the DNN models, thereby reducing storage and computation costs simultaneously. Existing work mainly focus on \emph{unstructured pruning} scheme~\cite{han2015learning,guo2016dynamic,liu2018rethinking} where arbitrary weight can be removed, and (coarse-grained) \emph{structured pruning} scheme~\cite{min20182pfpce,zhuang2018discrimination,zhu2018ijcai,ma2019tiny,zhao2019variational,Liu2020Autocompress} to eliminate whole filters/channels. The former results in high accuracy but limited hardware parallelism (and acceleration), while the latter is the opposite. Another active research area is the \textbf{Neural Architecture Search} (NAS) \cite{zoph2016neural}, which designs more efficient DNN architectures using automatic searching algorithms. EfficientNet \cite{tan2019efficientnet} and MobileNetV3 \cite{howard2019searching} are representative lightweight networks obtained by using NAS approaches. Recently, hardware-aware NAS \cite{tan2019mnasnet,wu2019fbnet,cai2018proxylessnas,jiang2019accuracy} has been investigated targeting acceleration on actual hardware platforms.

Different from the prior work on coarse-grained pruning and NAS that find a smaller, yet regular, DNN structure, recent work~\cite{ma2019pconv,niu2020patdnn,dong2020rtmobile} propose to prune the weights in a more fine-grained manner, e.g., assigning potentially different patterns to kernels. Higher accuracy can be achieved as a result of the intra-kernel flexibility, while high hardware parallelism (and mobile inference acceleration) can be achieved with the assist of compiler-level code generation techniques \cite{niu2020patdnn}. This work reveals a new dimension of optimization: \emph{With the aid of advanced compiler optimizations}, it is possible to achieve high accuracy and high acceleration simultaneously by injecting a proper degree of fine granularity in weight pruning. Despite the promising results, pattern-based pruning~\cite{ma2019pconv,niu2020patdnn} is only applied to 3$\times$3 convolutional (CONV) layers, which limits the applicability.

As the \textbf{first contribution}, we propose a general category of fine-grained structured pruning schemes that can be applied to various DNN layers, i.e., \emph{block-punched pruning} for CONV layers with different kernel sizes, and \emph{block-based pruning} for FC layers. We develop a comprehensive, compiler-based automatic code generation framework \emph{supporting the proposed pruning schemes in a unified manner}, \emph{supporting other types of pruning schemes}, and \emph{different schemes for different layers}. We show (i) the advantage of the proposed fine-grained structured pruning in both accuracy and mobile acceleration, and (ii) the superior end-to-end acceleration performance of our compiler framework on both dense (before pruning) and sparse DNN models.

While our compiler optimizations provide notable mobile acceleration and support of various sparsity schemes, it introduces \emph{a much larger model optimization space}: Different kernel sizes (1$\times$1, 3$\times$3, etc.) result in different acceleration performances under compiler optimizations, so do different sparsity schemes. Thus, it is desirable to perform a \emph{compiler aware, joint network pruning and architecture search}, determining the filter type and size, as well as pruning scheme and rate, for each individual layer. The \emph{objective} is to maximize accuracy satisfying a DNN latency constraint on the target mobile device. The DNN latency will be actually measured on the target mobile device, thanks to the fast auto-tuning capability of our compiler for efficient inference on different mobile devices.

We develop the compiler-aware NPAS framework to fulfill the above goal. It consists of three phases: (1) \emph{replacement of mobile-unfriendly operations}, (2) \emph{the core search process}, and (3) \emph{pruning algorithm search}. The overall latency constraint is satisfied through the synergic efforts of (i) incorporating the overall DNN latency constraint into the automatic search in Phase 2, and (ii) the effective search of pruning algorithm and performing weight training/pruning accordingly. As Phase 2 exhibits a larger search space than prior NAS work, to perform efficient search, we propose a meta-modeling procedure based on reinforcement learning (RL) with fast evaluation and Bayesian optimization. This will ensure the total number of training epochs comparable with representative NAS frameworks.

Our key contributions include: 
\begin{itemize}
    \item We propose a general category of fine-grained structured pruning applicable to various DNN layers, and a comprehensive, compiler code generation framework supporting different pruning schemes. We bridge the gap between model compression and NAS.
    \item We develop a compiler-aware framework of joint network pruning and architecture search, maximizing accuracy while satisfying inference latency constraint. 
    \item We design a systematic search acceleration strategy, integrating pre-trained starting points, fast accuracy and latency evaluations, and Bayesian optimization. 
    \item Our NPAS framework achieves by far the best mobile acceleration: 6.7ms, 5.9ms, and 3.9ms ImageNet inference times with 78.2\%, 75\%, and 71\% Top-1 accuracy, respectively, on an off-the-shelf mobile phone.
\end{itemize}

\section{Related Works} 
\label{sec:related_work}
\subsection{Network Pruning}
\label{sec:RW_pruning}

\begin{figure*}[t]
    \centering
    \includegraphics[width=0.9 \textwidth]{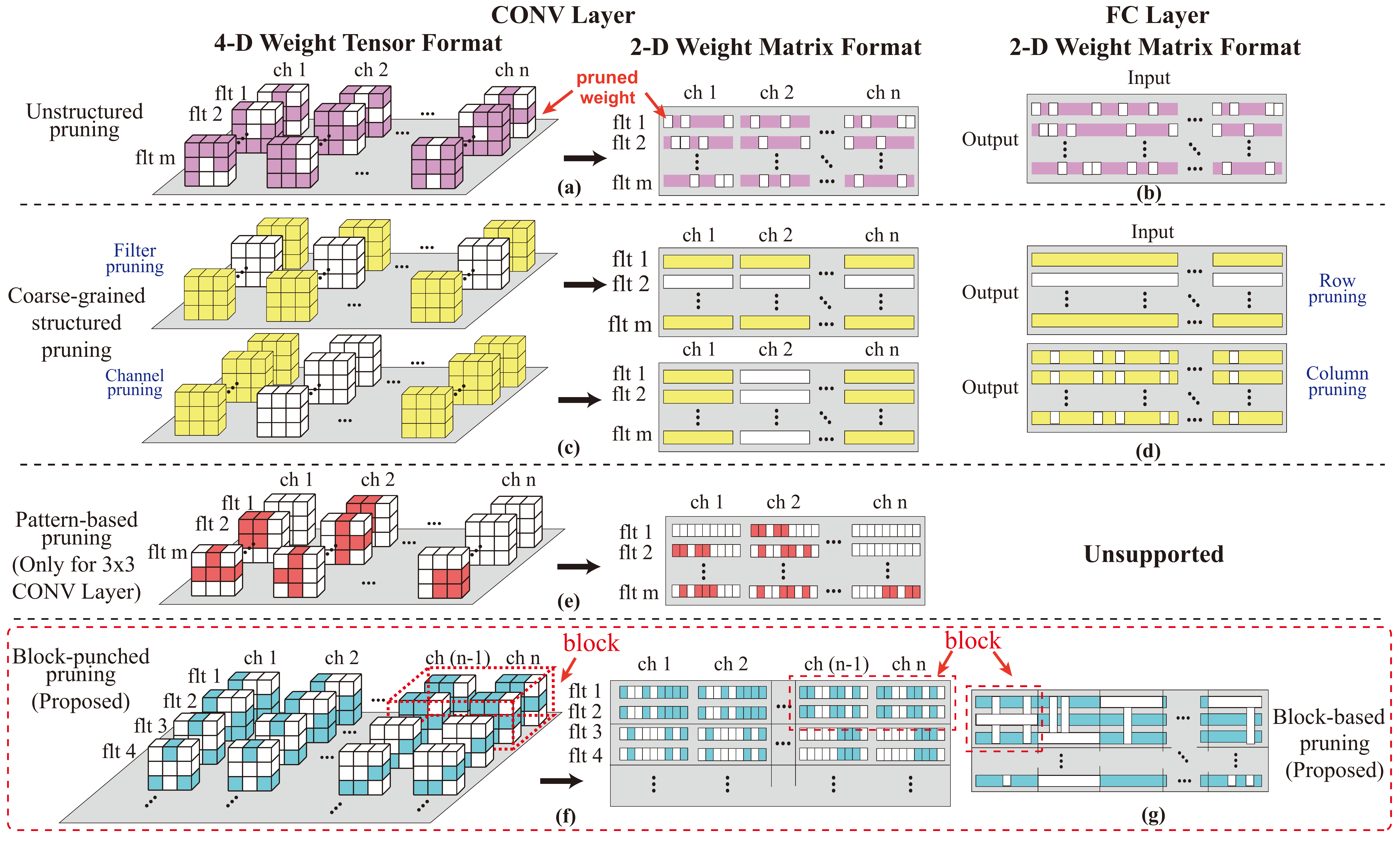}
    \caption{{Different weight pruning schemes for CONV and FC layers using 4D tensor and 2D matrix representation.}}
    \label{fig:prune_type}
\end{figure*}

Existing weight pruning research can be categorized according to pruning schemes and pruning algorithms.

\textbf{Pruning Scheme:}
Previous weight pruning work can be categorized into multiple major groups according to the pruning scheme: \textit{unstructured pruning}~\cite{han2015learning,guo2016dynamic,mao2017exploring}, \textit{coarse-grained structured pruning}~\cite{wen2016learning,he2017channel,luo2017thinet,yu2018nisp,liu2018rethinking,he2018soft,zhang2018systematic,li2019compressing,dong2019network}, and \textit{pattern-based pruning}~\cite{ma2019pconv,niu2020patdnn,ma2020image}.

Unstructured pruning (Fig.~\ref{fig:prune_type} (a) and (b)) removes weights at arbitrary position. Though it can significantly decrease the number of weights in DNN model as a fine-grained pruning scheme, the resulted sparse and irregular weight matrix with indices damages the parallel implementations and results in limited acceleration on hardware.

To overcome the limitation in unstructured, irregular weight pruning, many work~\cite{wen2016learning,he2017channel,liu2018rethinking,he2018soft,zhang2018systematic,li2019compressing,dong2019network,luo2017thinet,yu2018nisp,liu2019autocompress} studied the coarse-grained structured pruning at the level of filters and channels as shown in Fig.~\ref{fig:prune_type} (c) and (d). With the elimination of filters or channels, the pruned model still maintains the network structure with high regularity which can be parallelized on hardware. The downside of coarse-grained structured pruning is the obvious accuracy degradation by removing the whole filters/channels, which limits model compression rate.

Fig.~\ref{fig:prune_type} (e) shows the pattern-based pruning~\cite{ma2019pconv,niu2020patdnn,ma2020image} as a representative fine-grained structured pruning scheme. It assigns a pattern (from a predefined library) to each CONV kernel, maintaining a fixed number of weights in each kernel. As shown in the figure, each kernel reserves 4 non-zero weights (on a pattern) out of the original 3$\times$3 kernels. Besides being assigned a pattern, a kernel can be completely removed to achieve higher compression rate. Pattern-based pruning can simultaneously achieve high accuracy (thanks to the structural flexibility) and high inference acceleration with the aid of compiler-based executable code generation. Note that \textbf{compiler support} \cite{niu2020patdnn} is necessary for pattern-based pruning to deliver its promise on mobile acceleration. 

A limitation is that pattern-based pruning is limited to 3$\times$3 CONV layers in current work: 5$\times$5 or larger kernel size results in a large number of pattern types, which incurs notable computation overheads in compiler-generated executable codes. 1$\times$1 CONV layers and FC layers leave no space of designing different patterns for a kernel.

\textbf{Pruning Algorithm:}
Two main categories exist: \emph{heuristic pruning algorithm}~\cite{han2015deep,guo2016dynamic,dong2019network,luo2017thinet,yu2018nisp} and \emph{regularization-based pruning algorithm}~\cite{yuan2006model,wen2016learning,liu2018rethinking,he2017channel,he2018soft,zhang2018systematic,li2019compressing,he2019filter}.
Heuristic pruning was firstly performed in an iterative, magnitude-based manner on unstructured pruning~\cite{han2015deep}, and gets improved in later work~\cite{guo2016dynamic}. Heuristic pruning has also been incorporated into coarse-grained structured pruning~\cite{luo2017thinet,yu2018nisp,dong2019network}. 

Regularization-based algorithm uses mathematics-oriented method to deal with the pruning problem. Early work~\cite{wen2016learning,he2017channel} incorporates $\ell_1$ or $\ell_2$ regularization in loss function to solve filter/channel pruning problems. Later work~\cite{he2018soft} makes the regularization penalty ``softer" which allows the pruned filters to be updated during the training procedure. In~\cite{zhang2018systematic,li2019compressing}, an advanced optimization solution framework ADMM (Alternating Direction Methods of Multipliers) is utilized to achieve dynamic regularization penalty which significantly
reduces accuracy loss. In~\cite{he2019filter}, Geometric Median is proposed to conduct filter pruning.

\subsection{Neural Architecture Search (NAS)}

In general, NAS can be classified into the following categories by its searching strategy.
Reinforcement Learning (RL) methods \cite{zoph2016neural, zhong2018practical, zoph2018learning, baker2016designing,  liu2018progressive, cai2017efficient, pham2018efficient} employ Recurrent Neural Network (RNN) as predictor, with parameters updated by the accuracy of child network validated over a proxy dataset. 
Evolution methods \cite{real2019regularized, elsken2018efficient, real2019aging, miikkulainen2019evolving, xie2017genetic, liu2017hierarchical, xie2017genetic} develop a pipeline of parent initialization, population updating, generation and elimination of offsprings. 
One-shot NAS \cite{brock2017smash, bender2018understanding, you2020greedynas, guo2020single, chu2019fairnas} trains a large one-shot model containing all operations and shares the weight parameters to all candidate models. 
Gradient-based methods \cite{liu2018darts, cai2018proxylessnas, chen2020progressive, xu2019pc, wu2019fbnet, chu2019fair, fang2020densely} propose a differentiable algorithm distinct from prior discrete search, reducing searching cost while still getting comparable results. 
Bayesian optimization \cite{bergstra2013making, domhan2015speeding, mendoza2016towards, kandasamy2018neural, ru2020neural, whitedeep} uses optimal transport program to compute the distance of network architectures. 

Some recent work realize the importance of hardware co-design and incorporate the inference latency into NAS, which is more accurate than the intuitive volume estimation like Multiply–Accumulate operations (MACs) \cite{tan2019mnasnet, wu2019fbnet, cai2018proxylessnas}. MnasNet \cite{tan2019mnasnet} utilizes latency on mobile device as the reward to perform RL search, where gradient-based NAS work FBNet \cite{wu2019fbnet} and ProxylessNAS \cite{cai2018proxylessnas} add a latency term to the loss function. 
However, none of these hardware-targeting work fully exploit the potential of compiler optimizations or satisfy an overall latency requirement, not to mention accounting for compiler-supported sparse models. This motivates us to investigate another dimension of model optimization, that is, compiler-aware, latency-constrained, architecture and pruning co-search. 

\subsection{Compiler-assisted DNN Frameworks on Mobile}
Recently, mobile-based, compiler-assisted DNN execution frameworks~\cite{lane2016deepx,lane2015deepear,xu2018deepcache,huynh2017deepmon,yao2017deepsense,han2016mcdnn} have drawn broad attention from both industry and academia.
TensorFlow-Lite (TFLite)~\cite{TensorFlow-Lite}, Alibaba Mobile Neural Network (MNN)~\cite{Ali-MNN}, and TVM~\cite{chen2018tvm} are representative state-of-the-art DNN inference frameworks. Various optimization techniques, such as varied computation graph optimizations and half-float support, have been employed to accelerate the DNN inference on mobile devices (mobile CPU and GPU).

Recent work PatDNN~\cite{niu2020patdnn} and PCONV~\cite{ma2019pconv} employ a set of compiler-based optimizations to support specific pattern-based sparse DNN models to accelerate the end-to-end inference on mobile devices. 
However, the lack of support for different types of layers (e.g., 1$\times$1 CONV, 5$\times$5 CONV, and FC) limits the versatility of such framework.

\section{Proposed Fine-Grained Structured Pruning} 

Pattern-based pruning scheme~\cite{ma2019pconv,niu2020patdnn,ma2020image}, as mentioned in Section~\ref{sec:RW_pruning}, reveals a new optimization dimension of fine-grained structured pruning that can achieve high accuracy and high inference acceleration simultaneously with the assist of compiler optimizations. As pattern-based pruning is only applicable to $3\times3$ CONV layers, we propose a general category of fine-grained structured pruning scheme that can be applied to various DNN layers: block-based pruning for FC layers and block-punched pruning for CONV layers with different kernel sizes.

\textbf{Block-based Pruning:}
Fig.~\ref{fig:prune_type} (g) shows the block-based pruning scheme in 2D weight matrix format for FC layers. The entire weight matrix is divided into a number of equal-sized blocks, then the entire column(s) and/or row(s) of weights are pruned within each block.
Compared to the coarse-grained structured pruning, block-based pruning provides a finer pruning granularity to better preserve the DNN model accuracy. With an appropriate block size selected, the remaining computation within a block can still be parallelized on mobile device with the help of compiler. As a result, block-based pruning can achieve comparable hardware (inference) performance as coarse-grained structured pruning, under the same overall pruning rate.

\textbf{Block-punched Pruning:}
The CONV layers prefer the tensor-based computation rather than matrix-based computation used for FC layers. 
Inspired by block-based pruning, we develop block-punched pruning scheme tailored for CONV layers, which can be accelerated using the same compiler optimizations.
As shown in Fig.~\ref{fig:prune_type} (f), block-punched pruning requires pruning a group of weights at the same location of all filters and all channels within a block to leverage hardware parallelism from both memory and computation perspectives. With effective compiler-level executable code generation, high hardware parallelism (and inference acceleration on mobile) can also be achieved.

\textbf{Compiler Optimizations:} We develop a comprehensive, compiler-based automatic code generation framework supporting the proposed (block-punched/block-based) pruning schemes in a unified manner. It also supports other pruning schemes such as unstructured, coarse-grained, pattern-based pruning. In fact, unstructured and coarse-grained structured pruning schemes are just special cases of block-punched pruning, the former with block size $1\times1$ and the latter with block size of the whole weight tensor/matrix. A novel \emph{layer fusion} technique is developed, which is critical to the efficient implementation of super-deep networks. Fast \emph{auto-tuning} capability is incorporated for efficient end-to-end inference on different mobile CPU/GPU.

\begin{figure} [h]
     \centering
     \includegraphics[width=0.85\columnwidth]{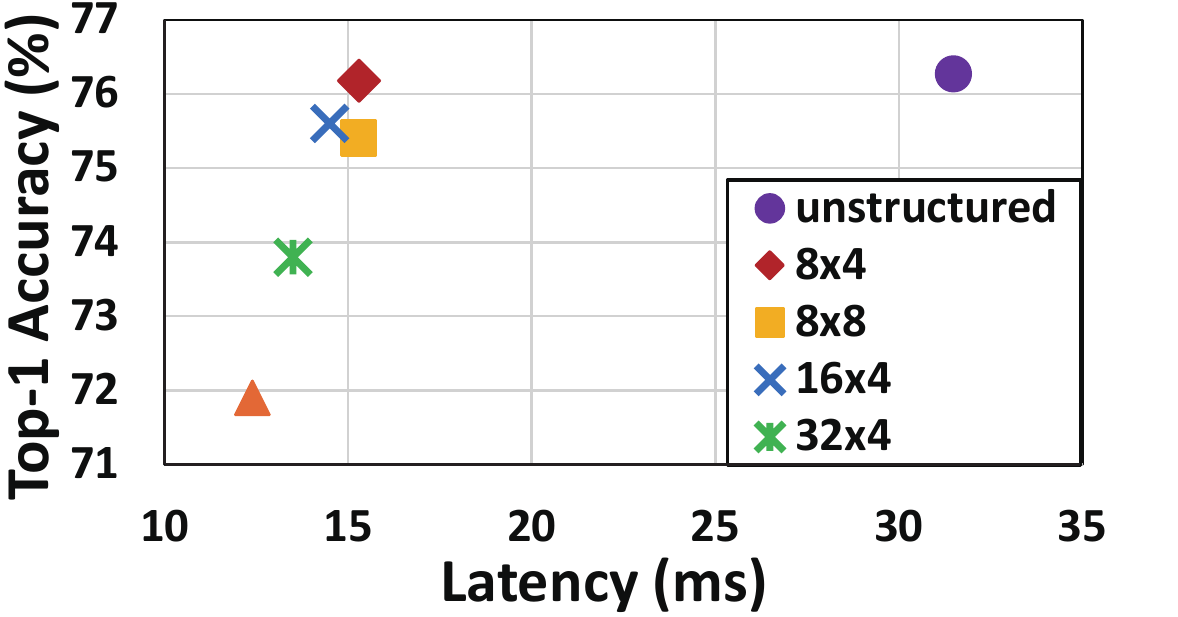} 
     \caption{Accuracy vs. Latency with different block sizes on ImageNet using ResNet-50 under uniform 6$\times$ pruning rate.}
     \label{fig:block_size_vs_latency}
\end{figure}

\textbf{Sample Results and Block Size Determination:}
Fig.~\ref{fig:block_size_vs_latency} shows example results of the accuracy vs. latency when applying block-punched pruning on ResNet-50 with different block sizes.
A uniform pruning rate (i.e., 6$\times$) and block size are adopted through all layers.
Under the same pruning rate, unstructured pruning (i.e., 1$\times$1 block size) preserves the highest accuracy but has the worst performance in latency. On the contrary, coarse-grained structured pruning (i.e., whole weight matrix as a block) achieves the lowest latency but with a severe accuracy degradation.
The results of block-punched pruning show high accuracy and high inference speed (low latency) simultaneously.

The reason is that the maximum hardware parallelism is limited by computation resources. Thus, even when dividing weights into blocks, each block's remaining weights are still sufficient to fulfill on-device hardware parallelism, especially on resource-limited mobile devices. 
One reasonable block size determination strategy is to let the number of channels contained in each block match the length of the vector register (e.g., 4) on target mobile CPU/GPU to ensure high parallelism. Then determine the number of filters to be contained (e.g., 8) by considering the given design targets.

\section{Motivation of Compiler-Aware Unified Optimization Framework}
\label{sec: motivation}
Our compiler optimizations provide notable acceleration of different filter types, and support for various sparsity schemes. A key \textbf{observation} is that different filter types and sparsity schemes have different acceleration performance under compiler optimizations (when computation (MACs) is the same). The following are measured on mobile CPU (Qualcomm Kryo 485) of a Samsung Galaxy S10 phone.

\begin{figure}[b!]
    \vspace{-3mm}
    \centering
    \includegraphics[width=1 \columnwidth]{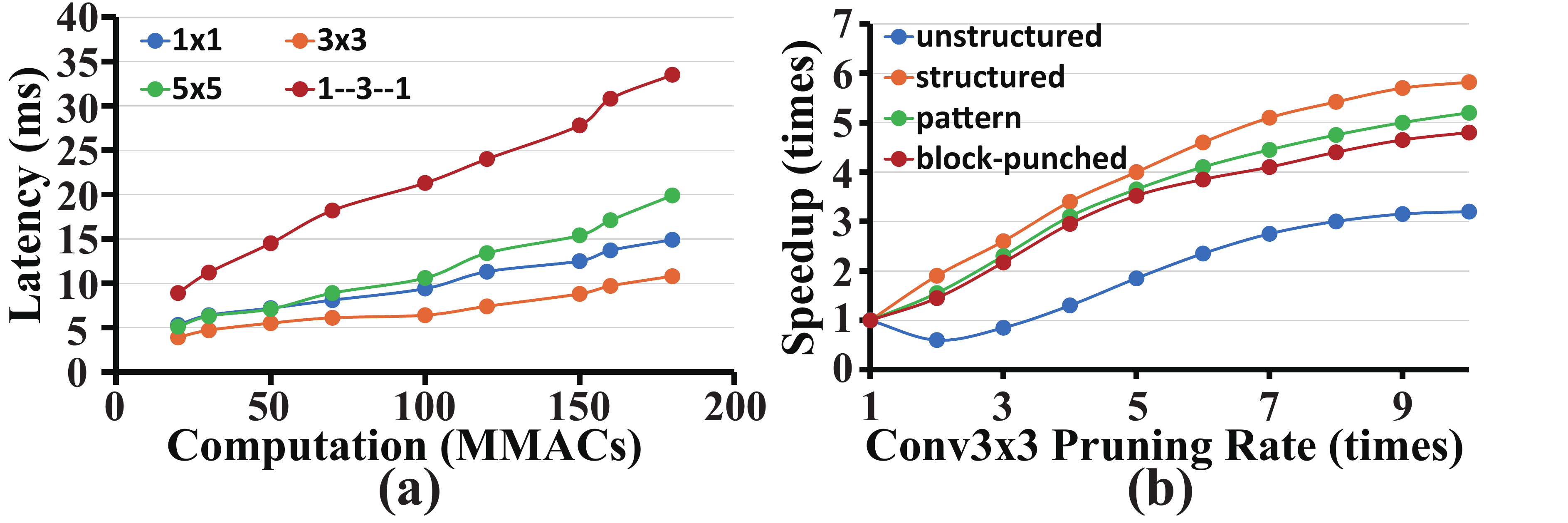}
    \vspace{-4mm}
    \caption{(a) Latency vs. Computation with different filter types, (b) speedup
    vs. pruning rate with different pruning schemes.}
    \label{fig:relation}
\end{figure}

\textbf{Different Filter Types (Kernel Sizes):}
Fig.~\ref{fig:relation} (a) shows the latency vs. computation (MACs) of a CONV layer with different kernel sizes. We fix the input feature map to 56$\times$56 and change the number of filters. Under the same computation, 3$\times$3 kernels achieve the best performance, where the 1$\times$1 kernels are the second.
Because 3$\times$3 kernels can be accelerated using Winograd algorithm, and makes it the most compiler-friendly; while 1$\times$1 kernels result in no input redundancy in GEMM computation, which also relieves the burden on compiler optimizations.

\textbf{Different Pruning Schemes:}
Fig.~\ref{fig:relation} (b) shows the computation speedup vs. pruning rate of a 3$\times$3 CONV layer with different pruning schemes. We choose the input feature map size of 56$\times$56 and 256 input and output channels.
We can observe that, with compiler optimizations, fine-grained pruning schemes (i.e., pattern-based and block-punched pruning) consistently outperform the unstructured pruning and achieve comparable acceleration compared to the coarse-grained structured pruning below 5$\times$ pruning.
Since, under reasonable pruning rate of fine-grained structured pruning schemes, the remaining weights in each layer are still sufficient to fully utilize hardware parallelism.

\textbf{Impact of Number of Layers:}
The number of computation layers is another critical factor that affects inference latency. 
To show the impact, we make a narrower-but-deeper version of ResNet-50 by doubling the number of layers, while keeping computation MACs the same as the original ResNet-50. 
And the inference speed of the narrower-but-deeper version is 1.22$\times$ slower than the original one using mobile GPU (44ms vs. 36ms).
The main reason is that a larger number of layers introduce more intermediate results and hence more frequent data access to the main memory. And the mobile CPU/GPU cannot be fully utilized due to a large number of memory-intensive layers.

Based on the above observations, it is desirable to perform a compiler-aware network pruning and architecture search, determining the \emph{filter type and size, as well as pruning scheme and rate} for each individual layer. 
The objective is to \emph{maximize DNN accuracy satisfying an inference latency constraint} when actually executing on the target mobile device, accounting for compiler optimizations.

\section{Proposed Unified Network Pruning and Architecture Search (NPAS) Algorithm} 

\subsection{Overview of NPAS Framework}

\begin{figure}[tb]
    \centering
    \includegraphics[width=1\columnwidth]{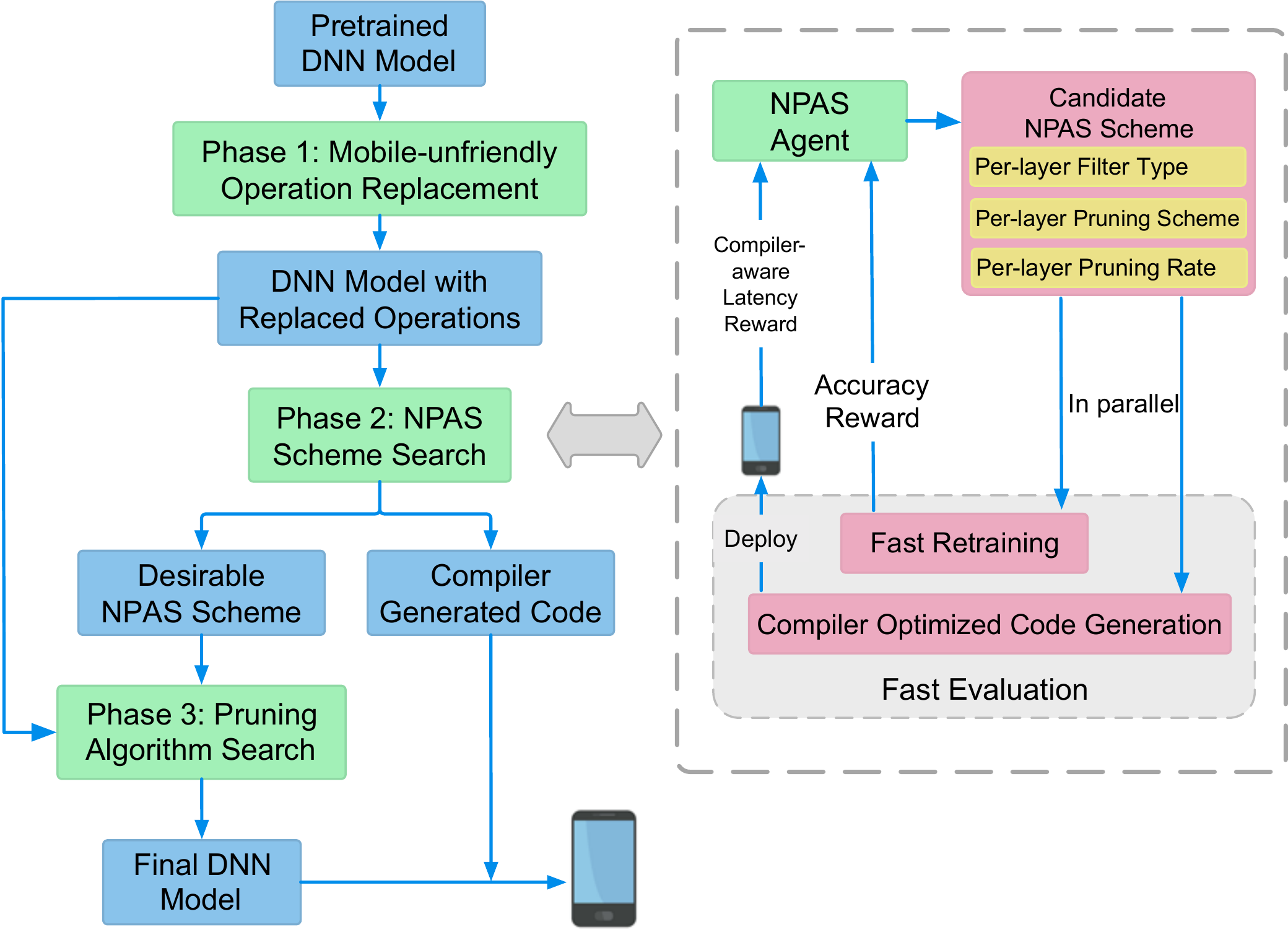}
    \caption{Overview of the proposed NPAS framework.}
    \label{fig:framework}
\end{figure}

Fig. \ref{fig:framework} shows the proposed NPAS framework. To take advantage of recent NAS results and accelerate the NPAS process, we start from a pre-trained DNN model, and go through three phases as shown in the figure.

\textbf{Phase 1: Replacement of Mobile-Unfriendly Operations:} Certain operators are inefficient to execute on mobile devices (mobile CPU and GPU). For instance, certain activation functions, such as sigmoid, swish, require exponential computation, and can become latency bottleneck on mobile inference. These unfriendly operations will be replaced by compiler-friendly alternatives such as hard-sigmoid and hard-swish, with negligible effect on accuracy.

\textbf{Phase 2: NPAS Scheme Search:} This phase generates and evaluates candidate \emph{NPAS schemes}, defined by the collection of per-layer filter types, per-layer pruning schemes and rates, and finally chooses the best-suited one. As per-layer pruning schemes and rates are being searched, Phase 2 exhibits \underline{a much larger search space than prior NAS}, which renders representative NAS algorithms like 
RL-based ones ineffective. To accelerate such search, we present \emph{a meta-modeling procedure based on RL with Bayesian Optimization} (BO), with details in Section~\ref{sec:details_phase2}. \emph{A fast accuracy evaluation method} is developed, tailored to NPAS framework.

Moreover, we incorporate the overall DNN latency constraint effectively in the reward function of NPAS scheme search, ensuring that such constraint can be satisfied at the search outcome. The overall DNN latency is actually measured on the target mobile CPU/GPU based on the candidate NPAS scheme currently under evaluation.
We rely on actual measurement instead of per-layer latency modeling as many prior NAS work. This is because our advanced compiler optimizations incorporate a strong layer fusion beyond prior compiler work, which is critical for efficient implementation of super-deep networks, and will make per-layer latency modeling less accurate.

\textbf{Phase 3: Pruning Algorithm Search:} The previous phase has already determined the per-layer pruning schemes and rates, so that the compiler-generated codes can satisfy the overall latency constraint. The remaining task of this phase is to search for the most desirable pruning algorithm to perform actual pruning and train the remaining weights\footnote{The above process cannot be accomplished by the fast accuracy evaluation in Phase 2 as we need to limit the number of training epochs.}. As the per-layer pruning rates are already determined, the candidate pruning algorithms to select from are limited to those with pre-defined per-layer pruning rates, including magnitude-based ones \cite{han2015deep, frankle2018lottery}, ADMM-based algorithm~\cite{zhang2018systematic,li2019compressing}, etc. As an extension over prior work, we generalize these algorithms to achieve different sparsity schemes with the help of group-Lasso regularization \cite{Kim_2012,wen2016learning}
In Phase 3, we compare the resulted DNN accuracy from the candidate pruning algorithms in a few epochs, select the one with the highest accuracy, and continue a best-effort algorithm execution to derive the final DNN model and compiled codes.

\subsection{Details of Phase 2: NPAS Scheme Search}
\label{sec:details_phase2}
\subsubsection{Search Space of NPAS in Phase 2}

\begin{table}[tbh]
\caption{NPAS search space for each DNN layer}
\small
\centering
\begin{threeparttable}
\scalebox{0.9}{
\begin{tabular}{c | c }
    \toprule
    \multirow{2}{*}{\makecell{ Filter \\ type}}     &       \{1$\times$1,\, 3$\times$3,\,  3$\times$3 DW \& 1$\times$1, \\
    &  1$\times$1 \& 3$\times$3 DW \& 1$\times$1,\, skipping\} \tnote{1} \\
  \midrule
    \multirow{2}{*}{\makecell{ Pruning \\ scheme}}     &       \{Filter \cite{zhuang2018discrimination}, 
      Pattern-based \cite{niu2020patdnn}, \\
     &  Block-punched/block-based\}  \\
\midrule
     \makecell{  Pruning   rate}   &    \{ 1$\times$, 2$\times$, 2.5$\times$, 3$\times$, 5$\times$, 7$\times$, 10$\times$ \}                  \\ 
      \bottomrule
\end{tabular}}
\end{threeparttable}
\begin{tablenotes}
\centering
\item[1] \small \& denotes cascade connection.
\end{tablenotes}
\label{Table: search_space}
\end{table}

\textbf{Per-layer filter types:}
As different filter types (kernel sizes) have different acceleration performance under compiler optimizations, the NPAS search space includes replacing the original filter type with 1$\times $1, 3$\times $3, a cascade of 3$\times$3 depth-wise (DW) and 1$\times$1 convolutions, a cascade of 1$\times$1 and 3$\times$3 DW and 1$\times$1 convolutions, or directly skipping the entire layer. The first two are most preferable with compiler optimizations (please refer to Section~\ref{sec: motivation}), and the cascade connection is shown in prior work \cite{howard2017mobilenets,sandler2018mobilenetv2} to provide the same accuracy with less computation.

\textbf{Per-layer pruning schemes:}
The NPAS agent can choose from filter (channel) pruning \cite{zhuang2018discrimination}, pattern-based pruning \cite{niu2020patdnn} and block-punched/based pruning for each layer. As different layers may have different compatible pruning schemes, we allow the NPAS the flexibility to choose different pruning schemes for different layers. This is well supported by our compiler code generation.  

\textbf{Per-layer pruning rate:}
We can choose from the list $\{ 1\times, 2\times, 2.5\times, 3\times, 5\times,  7\times, 10\times\}$ ($1\times$ means no pruning). 

\subsubsection{Q-Learning Training Procedure}
\label{sec:Q_learning}
As per-layer pruning scheme and rate is integrated in NPAS scheme search, the search space is beyond that of conventional NAS. To ensure fast search, we employ the RL algorithm Q-learning as the base technique, assisted by fast evaluation (Section~\ref{sec:fast_evaluation}) and Bayesian optimization (BO) (Section~\ref{sec:Bayesian}) for search speedup. 
The Q-learning algorithm consists of an NPAS agent, states and a set of actions.

For the state of the $i$-th layer in a given DNN, it is defined as a tuple of filter type, pruning scheme, and pruning rate 
i.e., $\{filter\_type_i,\ pruning\_scheme_i,\ pruning\_rate_i\}$, and each can be selected from the corresponding search space.
We add the layer depth to the state space to constrict the action space such that the state-action graph is directed and acyclic (DAG). 

For \emph{action space}, we allow transitions for a state with layer depth $i$ to a state with layer depth $i + 1$, ensuring that there are no loops in the graph.
This constraint ensures that the state-action graph is always a DAG. 
When layer depth reaches the maximum layer depth, the transition terminates.  

Based on above-defined state $s \in S$ and action $a \in A$, we adopt Q-learning procedure \cite{watkins1989learning} to update Q-values. 
We specify final and intermediate rewards as follows: 
\begin{equation}
r_T = V - \alpha \cdot \mathrm{max}(0, h - H),\ \ \ 
r_t = \frac{r_T}{T},
\end{equation}
where $V$ is the validation accuracy of the model, $h$ is the model inference speed or latency (actually measured on a mobile device), and $H$ is the threshold for the latency requirement. Generally, $r_T$ is high when the model satisfies the real-time requirement ($h<H$) with high evaluation accuracy. Otherwise the final reward is small, especially when the latency requirement is violated. 
For the intermediate reward $r_t$ which is usually ignored by setting it to zero \cite{baker2016designing} as it cannot be explicitly measured, the reward shaping~\cite{ng1999policy} is employed as shown above to speed up the convergence. Setting $r_t = 0$ could make the Q-value of $s_T$ much larger than others in the early stage of training, leading to an early stop of searching for the agent.

We adopt the $\epsilon$-greedy strategy \cite{mnih2015human} to choose actions. 
In addition, as the exploration space is large, the \emph{experience replay} technique is adopted for faster convergence \cite{experience_replay}. 
\subsubsection{Fast Evaluation Methods}
\label{sec:fast_evaluation}

We develop and adopt multiple tailored acceleration strategies to facilitate fast evaluation in NPAS scheme search.

\textbf{Unidirectional Filter Type Replacement:}
The NPAS scheme search needs to satisfy a pre-defined DNN latency constraint.
Thus, we follow the principle of not increasing kernel size to search per-layer filter type, which can effectively reduce search space. For example, we will no longer search the filter type for 1$\times$1 layers in the original model.

\textbf{Weight Initialization for Filter Type Candidates:}
The weights of the filter type candidate operators in each layer can be pre-trained before NPAS scheme search (Phase 2) very quickly using reconstruction error, which can make them act similarly to the original operations. Thus, the accuracy evaluation process can be significantly accelerated. 

\textbf{One-shot Pruning and Early Stopping for Fast Accuracy Evaluation:}
During the accuracy evaluation process, we follow the pruning scheme and rate (for a specific layer) in a candidate NPAS scheme, and conduct a one-shot pruning (on the target layer) based on weight magnitude. This straightforward pruning will result in accuracy degradation. But after a couple of epochs of retraining, it can distinguish the relative accuracy of different NPAS schemes. 

\textbf{Overlapping Compiler Optimization and Accuracy Evaluation:}
We use compiler code generation and actual on-device latency measurement because of (i) higher accuracy than per-layer latency modeling due to layer fusion mechanism 
, and (ii) the fast auto-tuning capability of compiler to different mobile devices. Please note that the compiler code generation and latency measurement \emph{do not need the absolute weight values}. Compiler code generation is much faster than DNN training (even a single epoch), and can be performed in parallel with accuracy evaluation (as accurate weight values are not needed). As a result, it will not incur extra time consumption to NPAS.

\subsubsection{Bayesian Predictor for Reducing Evaluations}
\label{sec:Bayesian}
As performing evaluation on a large amount of sampled NPAS schemes is time-consuming, we build a predictor with BO \cite{snoek2012practical,klein2017fast,chen2018bayesian}. 
The NPAS agent generates a pool of NPAS schemes. 
We first use BO to select a small number of NPAS schemes with potentially high rewards from the pool. Then the selected NPAS schemes are evaluated to derive more accurate rewards. We reduce the evaluation of NPAS schemes with possibly weak performance, thereby reducing the overall scheme evaluation effort.

We  build a predictor combining Gaussian process (GP) with a Weisfeiler-Lehman subtree (WL) graph kernel \cite{morris2017glocalized,shervashidze2011weisfeiler}
to handle the graph-like NPAS schemes.
The WL kernel compares two directed graphs in iterations.  In the $m$-th WL iteration, it first obtains the histogram of graph features $\phi_{m}(s)$ and $\phi_{m}(s')$ for two graphs. Then it compares the two graphs with  $k_{\mathrm{base}}\bigl(\phi_{m}(s), \phi_{m}(s')\bigr)$ where $k_{\mathrm{base}}$ is a base kernel and we employ dot product here. 
The iterative  procedure stops until 
$m=M$  and resultant WL kernel is
\begin{equation}
    k^{M}_{\mathrm{WL}}(s, s') = \sum_{m=0}^M w_m k_{\mathrm{base}}\bigl(\phi_{m}(s), \phi_{m}(s')\bigr).
    \label{eq:wl}
\end{equation}
where $w_m$ contains the weights for each WL iteration $m$, which is set to equal for all $m$ following \cite{shervashidze2011weisfeiler}. 
The \emph{Expected Improvement} \cite{qin2017improving} is employed as the acquisition function in the work. Algorithm \ref{alg: qb} provides a summary.

\begin{algorithm}[tb]
        \small
	    \caption{Q-learning with Bayesian Predictor Algorithm }\label{alg: qb}
	\begin{algorithmic}
		\STATE {\bf Input:}  Observation data $\mathcal{D}$, BO batch size $B$, BO acquisition function $\alpha(\cdot)$
		\STATE {\bfseries Output:} The best NPAS scheme $s$
        \FOR{steps}
        		\STATE Generate a pool of candidate NPAS schemes $ \mathcal{S}_c $;
        		\STATE Select $\{ \hat{s}^{ i} \}_{i=1}^B = \arg\max_{s \in  \mathcal{S}_c } \alpha (s \vert \mathcal{D})$;
        		\STATE Evaluate the scheme and obtain reward  $\{r^{ i}\}_{i=1}^B$ of $\{ \hat{s}^{ i} \}_{i=1}^B$;
        		\STATE Update Q values based on Q-learning with  reward;
        		\STATE $\mathcal{D}\leftarrow \mathcal{D} \cup (\{ s^{  i} \}_{i=1}^B , \{r^{i}\}_{i=1}^B)$;
        		\STATE Update GP of BO with  $\mathcal{D}$;
    	\ENDFOR
	\end{algorithmic}
\end{algorithm}

\begin{table*}[]
\caption{Comparison results of NPAS and representative lightweight networks.}
\centering
\footnotesize
\begin{tabular}{ccccccc}
\toprule
\multicolumn{1}{l}{} & A. / P. Search & Params & CONV MACs & Accuracy (Top-1/5) & Latency (CPU/GPU)   &  Device      \\
\hline
MobileNet-V1 \cite{howard2017mobilenets}         & N./N.          & 4.2M & 575M     & 70.6 / 89.5         & - / -    & -                 \\
MobileNet-V2 \cite{sandler2018mobilenetv2}         & N./N.          & 3.4M & 300M     & 72.0 / 91.0         & - / -    & -    \\
MobileNet-V3 \cite{howard2019searching}         & Y./N.          & 5.4M & 227M     & 75.2 / 92.2         & - / -    &  -    \\
NAS-Net-A \cite{zoph2018learning}            & Y./N.          & 5.3M & 564M     & 74.0 / 91.3         & 183ms / NA   & Google Pixel 1   \\
AmoebaNet-A \cite{real2019regularized}         & Y./N.          & 5.1M & 555M     & 74.5 / 92.0         & 190ms / NA    & Google Pixel 1  \\
MnasNet-A1 \cite{tan2019mnasnet}          & Y./N.          & 3.9M & 312M     & 75.2 / 92.5         & 78ms / NA      & Google Pixel 1  \\
ProxylessNas-R \cite{cai2018proxylessnas}       & Y./N.          &  NA   &     NA     & 74.6 / 92.2         & 78ms / NA      & Google Pixel 1  \\
\hline
NPAS (ours)  & Y./N.          & 5.3M & 385M     & 78.2 / 93.9         & 11.8ms / 6.7ms              & Galaxy S10  \\
NPAS (ours)        & Y./Y.          & 3.5M & 201M     & 75.0 / 92.0         & 9.8ms / 5.9ms          & Galaxy S10      \\
NPAS (ours)         & Y./Y.          & 3.0M & 147M     & 70.9 / 90.5              & 6.9ms / 3.9ms          & Galaxy S10    \\
NPAS (ours)         & Y./Y.          & 2.8M & 98M     &  68.3 / 89.4               & 5.6ms / 3.3ms         & Galaxy S10      \\
\bottomrule
\end{tabular}
\label{Table: comparison_with_sota}
\end{table*}

\section{Results and Evaluation} 

\subsection{Experimental Setup}

In this section, we use the image classification task and ImageNet dataset \cite{deng2009imagenet} to show the effectiveness of our framework.
All training processes use the SGD optimizer with a 0.9 momentum rate and a 0.0005 weight decay and use the batch size of 2048 per node. The starting learning rate is set to 0.001, and the cosine learning rate scheduler is used if not specified in our paper.
For Phase 1, we conduct a fast fine-tuning with 5 training epochs after replacing the mobile-unfriendly operations (only once for the entire NPAS process).
In Phase 2, 40 Nvidia Titan RTX GPUs are used to conduct the fast accuracy evaluation for candidate NPAS schemes concurrently. 
Since we start from a well-trained model, we retrain 2 epochs for each candidate one-shot pruned model for fast evaluation.
For each candidate model, we measure 100 runs of inference on target mobile devices and use the average value as end-to-end latency.
In Phase 3, we search the most desirable pruning algorithm including magnitude-based algorithm, ADMM-based algorithm~\cite{zhang2018systematic,li2019compressing} and geometric median-based algorithm~\cite{he2019filter} (only for filter pruning). We adopt 100 epochs for weight pruning and 100 epochs on remaining weights fine-tuning with knowledge distillation~\cite{shen2020meal}. 

The overall GPU days are varied based on pre-trained network and are reduced thanks to our fast evaluation and BO. 
For example, using EfficientNet-B0 as starting point, the overall searching time is 15 days, where Phase 1 only takes 5 epochs, and Phase 3 takes 1.5 days.

\subsection{Evaluation Results} 

\begin{figure}[h!]
    \centering
    \includegraphics[width=0.45 \textwidth]{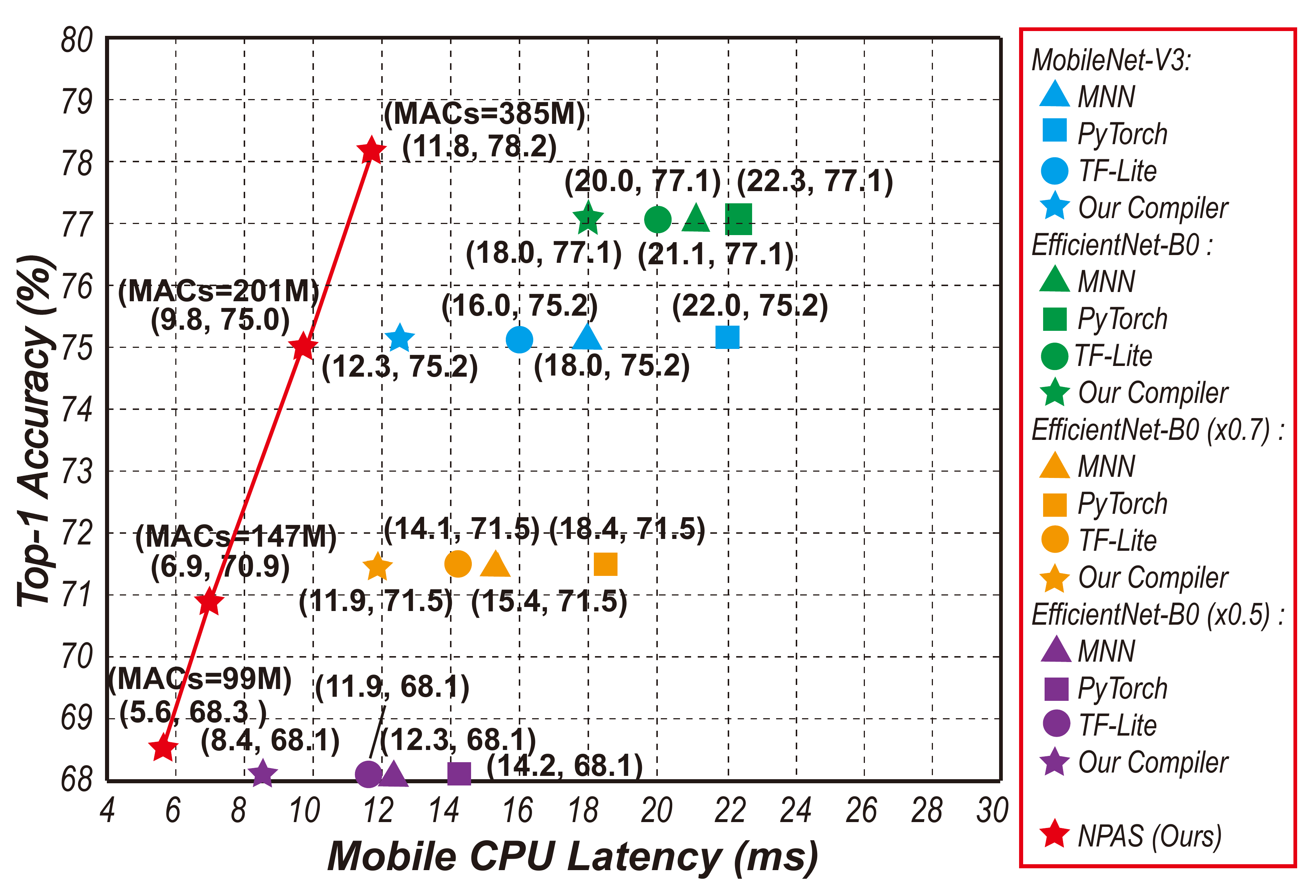}
    \caption{Accuracy vs. Latency comparison on mobile CPU.}
    \label{fig:accuracy_cpu}
\end{figure}

\begin{figure}[h!]
    \centering
    \includegraphics[width=0.45 \textwidth]{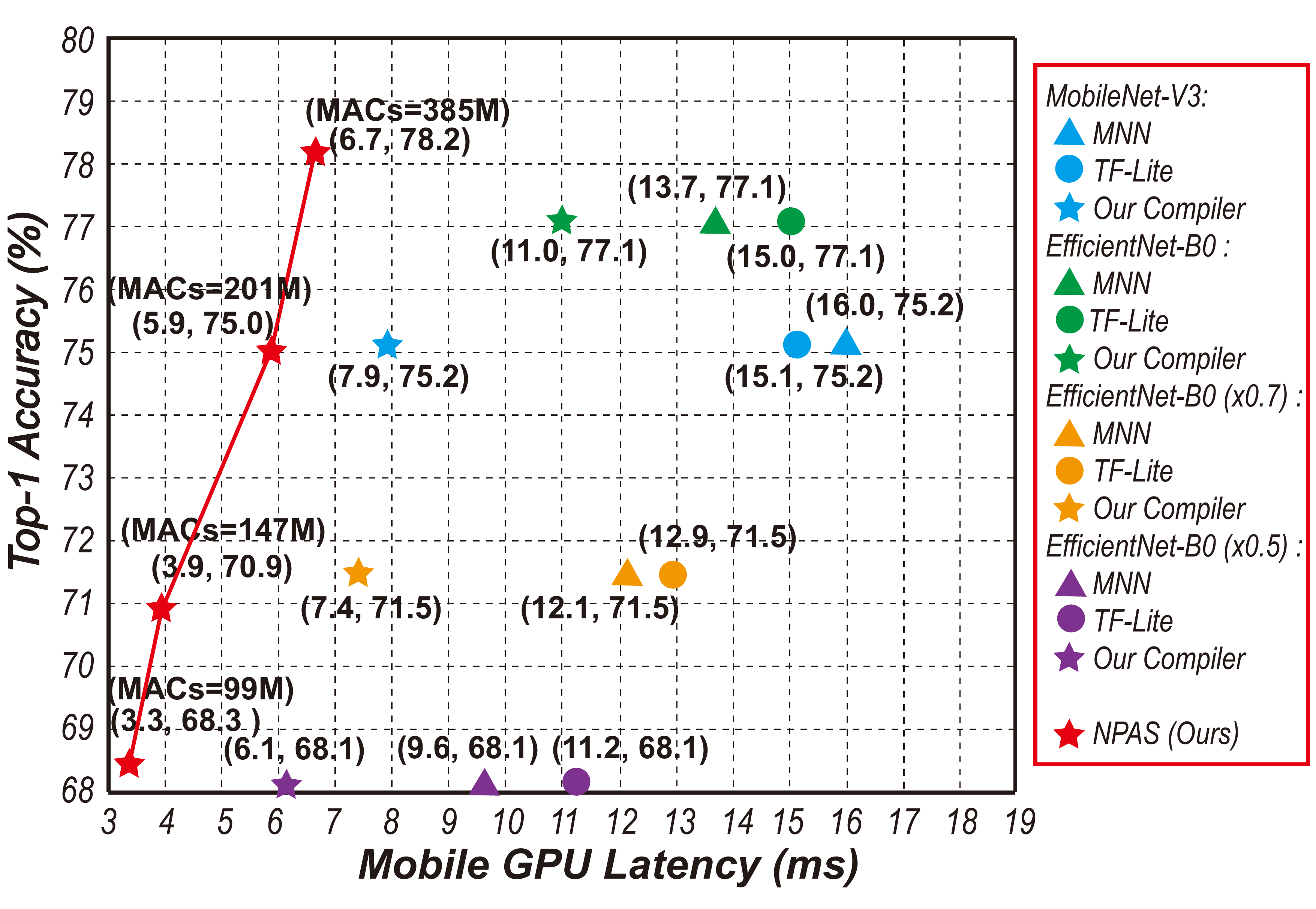}
    \caption{Accuracy vs. Latency comparison on mobile GPU.}
    \label{fig:accuracy_gpu}
\end{figure}

In Fig.~\ref{fig:accuracy_cpu} and \ref{fig:accuracy_gpu}, we compare our accuracy and latency results with representative DNN inference acceleration framework MNN, PyTorch Mobile, and TFLite.
Four dense DNN models are used for the comparisons, which are MobileNet-V3, EfficientNet-B0, shrunk versions of EfficientNet-B0 to 70\% original computation and 50\% original computation. 
The results are tested on a Samsung Galaxy S10 smartphone using mobile CPU (Qualcomm Kryo 485) or mobile GPU (Qualcomm Adreno 640). PyTorch Mobile does not support mobile GPU, so no corresponding results.
EfficientNet-B0 is used as our pretrained model. 

First, without incorporating NPAS, one can observe that our compiler optimizations can effectively speed up the same DNN inference, up to 46\% and 141\% (on MobileNet-V3), compared with the currently best framework MNN on mobile CPU and GPU, respectively. The red star shapes in the figures represent the NPAS generated results under different latency constraints. Our NPAS results consistently outperform the representative DNN models, and achieve the Pareto optimality in terms of accuracy and inference latency. For the starting models that have already met the latency constraint, we replace the mobile-unfriendly operations and maintain the original architecture.
With MobileNet-V3 level accuracy (75\% Top-1), our inference time (201M MACs) is 9.8ms and 5.9ms, respectively. With MobileNet-V2 level accuracy (71\% Top-1), the inference time of NPAS solution (147M MACs) is 6.9ms and 3.9ms, respectively. To the best of our knowledge, this is never accomplished by any existing NAS or weight pruning work.

Table~\ref{Table: comparison_with_sota} shows the model details, with representative handcrafted and hardware-aware NAS models as references. One can observe the computation (MACs) reduction under the same accuracy compared with the prior references, thanks to the joint network pruning and search. One can also observe the huge gap in latency compared with these prior work, as neither of compiler optimizations nor compiler-aware optimizations are accounted for. 
This gap is the reason we believe that compiler optimizations and awareness will contribute significantly to DNN accelerations.

\section{Conclusion} 
In this work, we propose (i) a fine-grained structured pruning applicable to various DNN layers, and (ii) a compiler automatic code generation framework supporting different DNNs and different pruning schemes, which bridge the gap of model compression and NAS. We further propose NPAS, a compiler-aware unified network pruning and architecture search, and several techniques are used to accelerate the searching process.

\newpage
\section{Acknowledgements}
This research is partially funded by National Science Foundation CCF-1901378, CCF-1919117, and CCF-1937500, Army Research Office/Army Research Laboratory via grant W911NF-20-1-0167 (YIP) to Northeastern University, a grant from Semiconductor Research Corporation (SRC), and a grant from Jeffress Trust Awards in Interdisciplinary Research.
Any opinions, findings, and conclusions or recommendations  in this material are those of the authors and do not necessarily reflect the views of NSF, ARO, SRC, or Thomas F. and Kate Miller Jeffress Memorial Trust.

{\small

\bibliographystyle{ieee_fullname}
}

\end{document}